\documentclass[10pt,twocolumn,letterpaper]{article}

\usepackage[pagenumbers]{wacv} 
\usepackage{graphicx}
\usepackage{amsmath}
\usepackage{amssymb}
\usepackage{booktabs}
\usepackage{multirow}
\usepackage{svg}
\usepackage{tabularx}
\usepackage{arydshln}
\usepackage{subcaption}

\usepackage[pagebackref,breaklinks,colorlinks]{hyperref}

\usepackage[capitalize]{cleveref}
\crefname{section}{Sec.}{Secs.}
\Crefname{section}{Section}{Sections}
\Crefname{table}{Table}{Tables}
\crefname{table}{Tab.}{Tabs.}

\def\wacvPaperID{1576}
\def\confName{WACV}
\def\confYear{2025}

\begin{document}

\title{CoVLA: Comprehensive Vision-Language-Action Dataset\\for Autonomous Driving}

\author{
    Hidehisa Arai$^{*}$ \quad Keita Miwa$^{*}$ \quad Kento Sasaki$^{*}$ \\
    Kohei Watanabe \quad Yu Yamaguchi \quad Shunsuke Aoki \quad Issei Yamamoto \\
    Turing Inc. \\
    {\tt\small \url{https://turingmotors.github.io/covla-ad/}}
}
\twocolumn[{
\renewcommand\twocolumn[1][]{#1}%
\maketitle
\begin{center}
    \centering
    \includegraphics[width=1.0\linewidth]{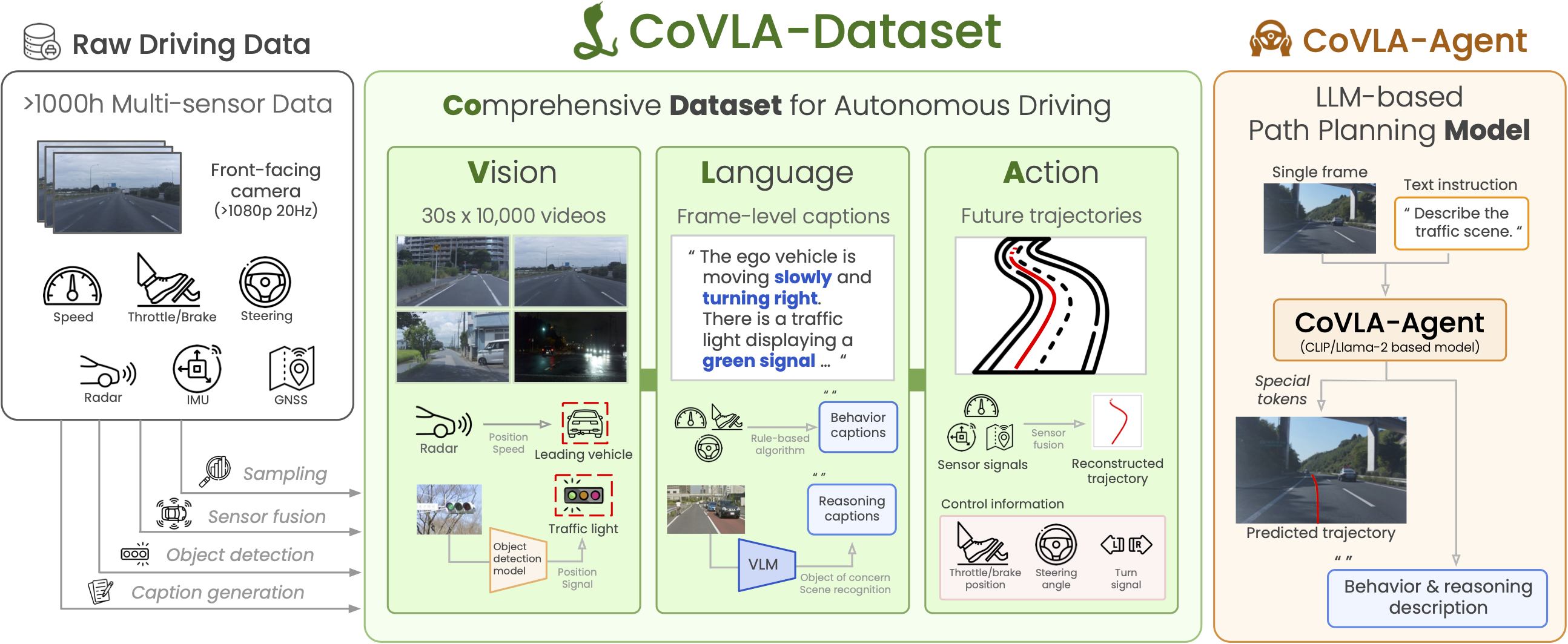}
    \captionof{figure}{\label{fig:teaser}{\textbf{CoVLA framework overview}. We develop \textbf{\textcolor[HTML]{006400}{CoVLA-Dataset}}, a comprehensive dataset for autonomous driving encompassing unique 10,000 video clips, frame-level language captions describing the driving scenarios, and future trajectory actions. We also show \textbf{\textcolor[HTML]{cc5500}{CoVLA-Agent}}, a VLM-based path planning model capable of predicting the future trajectory of the vehicle and providing a textual description of its behavior and reasoning.}
    \label{fig:overview}
    }
\end{center}}]

\renewcommand{\thefootnote}{\fnsymbol{footnote}}
\footnotetext[1]{\ equal contribution.}
\renewcommand{\thefootnote}{\arabic{footnote}}

\begin{abstract}
Autonomous driving, particularly navigating complex and unanticipated scenarios, demands sophisticated reasoning and planning capabilities. While Multi-modal Large Language Models (MLLMs) offer a promising avenue for this, their use has been largely confined to understanding complex environmental contexts or generating high-level driving commands, with few studies extending their application to end-to-end path planning. A major research bottleneck is the lack of large-scale annotated datasets encompassing vision, language, and action. To address this issue, we propose CoVLA (Comprehensive Vision-Language-Action) Dataset, an extensive dataset comprising real-world driving videos spanning more than 80 hours. This dataset leverages a novel, scalable approach based on automated data processing and a caption generation pipeline to generate accurate driving trajectories paired with detailed natural language descriptions of driving environments and maneuvers. This approach utilizes raw in-vehicle sensor data, allowing it to surpass existing datasets in scale and annotation richness. Using CoVLA, we investigate the driving capabilities of MLLMs that can handle vision, language, and action in a variety of driving scenarios. Our results illustrate the strong proficiency of our model in generating coherent language and action outputs, emphasizing the potential of Vision-Language-Action (VLA) models in the field of autonomous driving. This dataset establishes a framework for robust, interpretable, and data-driven autonomous driving systems by providing a comprehensive platform for training and evaluating VLA models, contributing to safer and more reliable self-driving vehicles. The dataset is released for academic purpose.
\end{abstract}

\section{Introduction}
\label{sec:intro}
A key challenge for autonomous driving technology lies in addressing the ``long tail'' of diverse and unpredictable driving environments~\cite{makansi2021exposing, zhou2022long}.
Self-driving vehicles must navigate not only common scenarios but also rare and complex situations, necessitating diverse world knowledge and advanced reasoning capabilities~\cite{fu2024drive}. This demands a deep understanding of the environment and a capacity for reasoning that extends beyond recognizing objects to interpreting their behavior and planning actions accordingly.
Vision-Language-Action (VLA) models, by seamlessly integrating visual perception with language understanding and action planning, have emerged as a promising pathway towards this goal. Recent advancements in VLA, particularly in robotics~\cite{rt22023arxiv,padalkar2023open,kim24openvla} and autonomous driving~\cite{renz2024carllava}, demonstrate their potential for enabling more robust and intelligent driving systems.

However, a major obstacle in applying VLA models to autonomous driving is the lack of large-scale datasets that effectively combine visual data with language descriptions and driving actions. Existing datasets often fall short in terms of scale and comprehensive annotation, especially for language, which often requires laborious manual effort. This limits the development and evaluation of robust VLA models capable of handling the complexities of real-world driving.

This paper introduces CoVLA (\textbf{Co}mprehensive \textbf{V}ision-\textbf{L}anguage-\textbf{A}ction) Dataset, a novel large-scale dataset designed to overcome these limitations. CoVLA-Dataset utilizes scalable automated approaches for labeling and captioning, creating a rich dataset of 10,000 real-world driving scenes, totaling more than 80 hours of videos. Each 30-second scene contains accurate driving paths and detailed descriptions in natural language, derived from synchronized front-facing camera footage and data from in-vehicle sensors. This rich dataset allows for a deeper understanding of the driving environment and agent behavior.
To demonstrate its effectiveness in advancing autonomous driving research, we develop CoVLA-Agent, a VLA model trained on our dataset for trajectory prediction and traffic scene description generation. Our findings indicate that our VLA model is capable of making consistent and precise predictions, even in situations that demand complex and advanced judgements.
The main contributions of this paper are summarized as follows:
\begin{itemize}
    \item We introduce CoVLA Dataset, a large-scale dataset that provides trajectory targets for diverse driving scenarios, as well as detailed frame-by-frame situational descriptions.
    \item We present a scalable method that accurately estimates trajectories through sensor fusion and automatically generates frame-level text captions of critical driving information.
    \item We develop CoVLA-Agent, a novel VLA model for interpretable end-to-end autonomous driving on top of CoVLA Dataset. Our model demonstrate the capability to consistently generate driving scene descriptions and predict trajectories, paving the way for more reliable autonomous driving. 
\end{itemize}

\section{Related Works}

\label{sec:related_works}

\begin{table*}[ht]
\setlength\dashlinedash{0.5pt}
\setlength\dashlinegap{2.0pt}

\small
\centering
\begin{tabular}{lcccccccc}
\toprule
\multirow{2}{*}{\textbf{Dataset}} & \multirow{2}{*}{\textbf{Year}} & \multicolumn{2}{c}{\textbf{Vision Data}} & \multicolumn{3}{c}{\textbf{Language Data}} &\multicolumn{2}{c}{\textbf{Action Data}}  \\
\cmidrule(lr){3-4}
\cmidrule(lr){5-7}
\cmidrule(lr){8-9}
& & \textbf{Frames} & \textbf{Source} & \textbf{Type} & \textbf{Method} & \textbf{Count} & \textbf{Type} & \textbf{Method} \\
\midrule
BDD-X~\cite{kim2018textual}              & 2018 & 26.2K & BDD~\cite{Yu_2020_CVPR} & Caption & Manual & 26K & --- & --- \\ 
\hdashline
BDD-OIA~\cite{xu2020bddoia}              & 2020 & 22.9K & BDD~\cite{Yu_2020_CVPR} & Caption & Manual & 27K & Command & Manual \\ 
\hdashline
\multirow{2}{*}{DRAMA~\cite{Malla_2023_WACV}} & \multirow{2}{*}{2023} & \multirow{2}{*}{17.8K} & Original & \multirow{2}{*}{Caption} & \multirow{2}{*}{Manual} & \multirow{2}{*}{17K} & \multirow{2}{*}{Command} & \multirow{2}{*}{Manual} \\ 
&&&(Real)\\
\hdashline
\multirow{2}{*}{OpenDV-2K~\cite{yang2024genad}} & \multirow{2}{*}{2024} & \multirow{2}{*}{65.1M} & YouTube & \multirow{2}{*}{Caption} & Auto & \multirow{2}{*}{65.1M} & \multirow{2}{*}{Command} & \multirow{2}{*}{Auto} \\
&&&Public Datasets && (Single)&&&\\\hdashline
HAD~\cite{kim2019CVPR}                   & 2019 & 1.1M & HDD~\cite{ramanishka2018CVPR} & Caption & Manual & 47K & Traj, Command & Base \\\hdashline
Talk2Car~\cite{deruyttere-etal-2019-talk2car} & 2019 & 400K & nuScenes~\cite{Caesar_2020_CVPR} & Caption & Manual & 12K & Traj & Base \\ \hdashline
Talk2Car-Trajectory~\cite{deruyttere2022talk2car} & 2022 & 400K & nuScenes~\cite{Caesar_2020_CVPR} & Caption & Manual & 12K & Traj, Command & Base, Manual \\ \hdashline
DriveLM-nuScenes~\cite{sima2023drivelm}  & 2023 & 4.8K & nuScenes~\cite{Caesar_2020_CVPR} & QA & Manual & 445K & Traj & Base \\ \hdashline
\multirow{2}{*}{DriveLM-CARLA~\cite{sima2023drivelm}}     & \multirow{2}{*}{2023} & \multirow{2}{*}{183K} & Original & \multirow{2}{*}{QA} & Auto & \multirow{2}{*}{3.76M} & \multirow{2}{*}{Traj} & \multirow{2}{*}{Sim} \\
&&&(Sim)&&(Single)\\
\hline
\multirow{2}{*}{\textbf{CoVLA (ours)}} & \multirow{2}{*}{2024} & \multirow{2}{*}{6M} & Original & \multirow{2}{*}{Caption}  & Auto  & \multirow{2}{*}{6M} & \multirow{2}{*}{Traj} & \multirow{2}{*}{GPS/IMU} \\
&&&(Real)&& (Multi) &&&\\
\bottomrule
\end{tabular}
\caption{\textbf{Comparison of driving datasets with language and action data}.}
\label{tab:comparison}
\end{table*}
\subsection{Datasets for Autonomous Driving}

Research in autonomous driving has been deeply connected with the development of large-scale datasets~\cite{li2024_driving_dataset_survey, liu2024survey}. The evolution of these datasets has seen a clear trend toward incorporating richer sensors and modalities, moving from a focus on individual tasks like object recognition to more complex challenges such as scene understanding, motion prediction, and end-to-end driving. Early efforts primarily focused on the recognition of traffic environments, proposing datasets for tasks such as pedestrian recognition~\cite{dalal2005histograms, dollar2011pedestrian, zhang2017citypersons, neumann2019nightowls}, traffic light and sign recognition~\cite{stallkamp2012man, houben2013detection, zhu2016traffic, zhang2022cctsdb, de2009real, philipsen2015traffic, behrendt2017deep, fregin2018driveu}, and driving scene segmentation~\cite{brostow2008segmentation, cordts2016cityscapes, neuhold2017mapillary, scharwachter2013efficient, yu2020bdd100k}. 

The increasing complexity of autonomous driving tasks led to the development of large-scale datasets that incorporate data from various sensors. KITTI~\cite{geiger2012we} is the pioneering work, which provides driving footage recorded from front-facing cameras and LiDAR. It also includes 3D bounding box annotations and ground truth trajectories. Waymo Open Dataset~\cite{sun2020scalability}, while employing a similar approach, has significantly enhanced both scale and diversity. Argoverse 1~\cite{chang2019argoverse}, Argoverse 2~\cite{wilson2023argoverse}, and nuScenes~\cite{caesar2020nuscenes} provide High-Definition Maps (HD-Maps) along with multi-sensor information and 3D bounding box annotations to make them more suitable for motion prediction tasks. Lyft L5~\cite{houston2021one} and nuPlan~\cite{caesar2021nuplan}, which also include HD-Maps, are more focused on planning tasks. These rich, multi-modal datasets have significantly advanced research in areas such as 3D object detection, trajectory prediction, and behavior planning, enabling the development of algorithms capable of handling more complex and realistic driving scenarios.

In recent years, the development of Large Language Models (LLMs) has given rise to attempts to employ Multi-modal LLMs (MLLMs) for complex driving environment understanding and driving operation planning. This has led to the emergence of Vision-Language datasets specifically tailored to traffic environments. BDD-X~\cite{kim2018textual} provides manually annotated free-form captions that describe the reasons behind the ego vehicle actions in driving videos. In BDD-OIA~\cite{xu2020bddoia} dataset, closed-form explanations about objects that trigger responses from the ego vehicle are provided, along with recommended actions. Similarly, DRAMA~\cite{Malla_2023_WACV} provides closed-form and free-form captions aimed at identifying risks during driving. It also includes annotations of objects related to risk events and recommended action commands. OpenDV-2K~\cite{yang2024genad} is a large-scale dataset derived from publicly available videos, annotated with automatically generated text captions and driving action commands. However, these datasets provide only high-level driving commands (e.g., ``stop'' and ``turn left'') as actions and often lack fine-grained trajectory information, which is essential for representing detailed driving maneuvers and training end-to-end autonomous driving systems.
HAD~\cite{kim2019CVPR}, Talk2Car~\cite{deruyttere-etal-2019-talk2car}, Talk2Car-Trajectory~\cite{deruyttere2022talk2car}, and DriveLM~\cite{sima2023drivelm} are Vision-Language datasets that include trajectory information as finer-grained action annotations.
However, these datasets are often limited in scale, particularly in terms of the diversity and complexity of the driving scenarios they cover, hindering their ability to support the development of truly robust and generalizable VLA models.

\subsection{MLLMs for Driving Tasks}

Addressing the complex and often unpredictable nature of real-world driving scenarios, particularly those falling within the ``long tail'' of the driving distribution (e.g., navigating road closures), necessitates a higher level of reasoning and decision-making capabilities in autonomous systems. Recent research has increasingly focused on incorporating MLLMs into autonomous driving systems, aiming to leverage their powerful reasoning and language understanding capabilities to address these challenges. For instance, ADAPT~\cite{jin2023adapt} utilizes Vision-Language Models (VLMs) to generate natural language explanations of driving actions and provide reasoning behind them, enhancing the interpretability of the system. Early attempts to predict driving actions using language models involved employing GPT-3.5 and GPT-4~\cite{achiam2023gpt} to predict high-level driving commands from situational descriptions~\cite{fu2024drive, wen2024dilu}. These initial efforts set the stage for more recent studies, which have explored fine-tuning LLMs to predict various types of driving behavior like trajectories~\cite{xu2023drivegpt4, sima2023drivelm} or control signals such as steering and acceleration~\cite{chen2023driving, mao2023gpt, sha2023languagempc} through textual outputs. While these approaches handle driving maneuver directly through language, LMDrive~\cite{shao2024lmdrive} and DriveMLM~\cite{wang2023drivemlm} predict control signals by passing the embeddings from the Transformer outputs to decoders. CarLLaVA~\cite{renz2024carllava} uses the LLaVA architecture to predict trajectories by taking images, vehicle speed, and target location coordinates as inputs.
Notably, this approach has achieved significantly superior performance compared to existing methods on the CARLA leaderboard 2.0, demonstrating the effectiveness of VLMs for complex trajectory prediction tasks.
\section{Dataset}
This section provides an in-depth look at CoVLA-Dataset, detailing its structure, content, and the methods employed to create this valuable resource for autonomous driving research. We highlight its coverage of diverse real-world driving scenarios, synchronized multi-modal data streams (front-facing camera, in-vehicle signals and other sensors) and its large-scale annotated data: 10,000 driving scenes totaling over 80 hours of video, each with accurate frame-level trajectory and descriptive caption annotations. To create this extensive VLA dataset, we develop a novel and scalable approach for automatically generating scene descriptions and ground truth trajectories from raw data.

\subsection{Development}
\subsubsection{Raw Data}

We develop multiple data collection vehicles to acquire real-world driving data, including sensor data from the front-facing camera, Controller Area Network (CAN) bus, Global Navigation Satellite System (GNSS), and Inertial Measurement Unit (IMU) and deploy them around Tokyo, Japan.

\begin{itemize}
    \item \textbf{Data collection period:} 6 months (summer to winter), resulting in over 1000 hours of raw data.
    \item \textbf{Data collection environments:} Various locations in and around Tokyo (urban centers, complex highway interchanges, narrow residential streets, mountainous areas with winding roads), different weather conditions (sunny, cloudy, rainy, heavy rain), and various times of day (daytime, evening, nighttime).
    \item \textbf{Equipment:} Front-facing camera (1928$\times$1208 pixels, 20 FPS, H.265 codec), CAN bus (accelerator/brake pedal position, steering wheel angle, turn signal status, gear position, vehicle speed), GNSS and IMU (binary format with timestamps), in-vehicle storage devices.
\end{itemize}

\subsubsection{Data Sampling}
\label{subsubsec:data_sampling}

From over 1,000 hours of raw driving data, we carefully select several hundred hours that meet the following criteria: (1) recorded while the vehicle is in driving gear, (2) the maximum speed does not exceed 100 km/h, and (3) GNSS data is continuously available.

To emphasize the diversity of driving scenarios, we sample individual data points with weighting inversely proportional to the pre-computed empirical distribution of a given set of features. Specifically, we select the maximum absolute value of steering angle, the maximum absolute value of acceleration, and turn signal as the features to balance over, applying binning for the first two. We calculate the empirical joint distribution of these categorical features. After applying additive smoothing~\cite{chen1999empirical} with a smoothing parameter of \( \delta=50 \), we use the inverse of these values as probability weights during sampling. Based on these adjustments, we select 10,000 diverse 30-second scenes through this method, resulting in a total of 6,000,000 video frames, equivalent to 83.3 hours of driving data.

\subsubsection{Auto Labeling}
\begin{figure}[t!]
    \centering
    \includegraphics[width=1.0\linewidth]{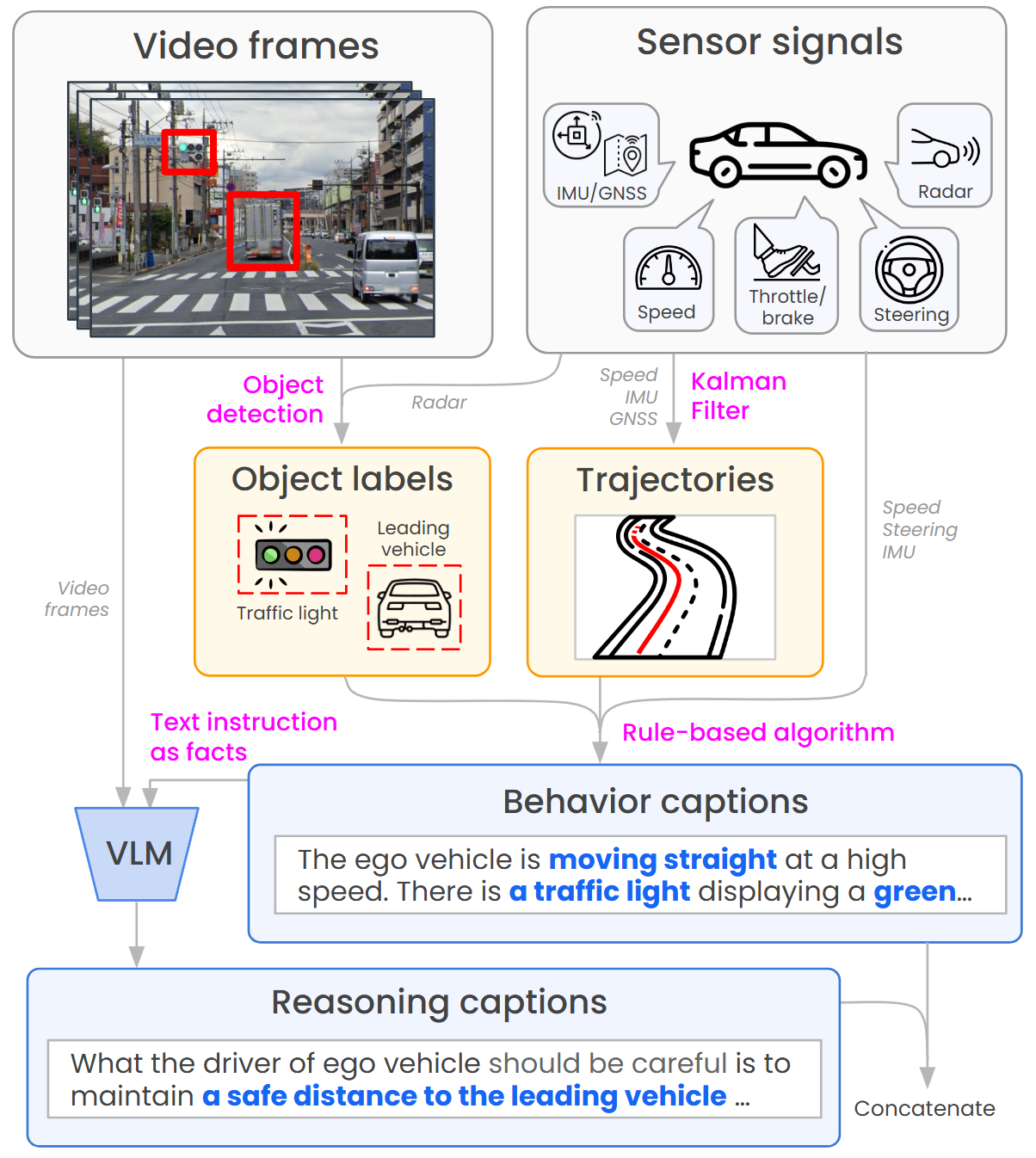}
    \caption{\textbf{Overview of the dataset generation pipeline}. We automatically label video frames and sensor signals to generate trajectories and other labels. Furthermore, we apply auto-captioning to the video frames to produce both behavior and reasoning captions.}
    \label{fig:dataset_creation}
\end{figure}
\begin{figure*}[t]
    \centering
    \includegraphics[width=1.0\linewidth]{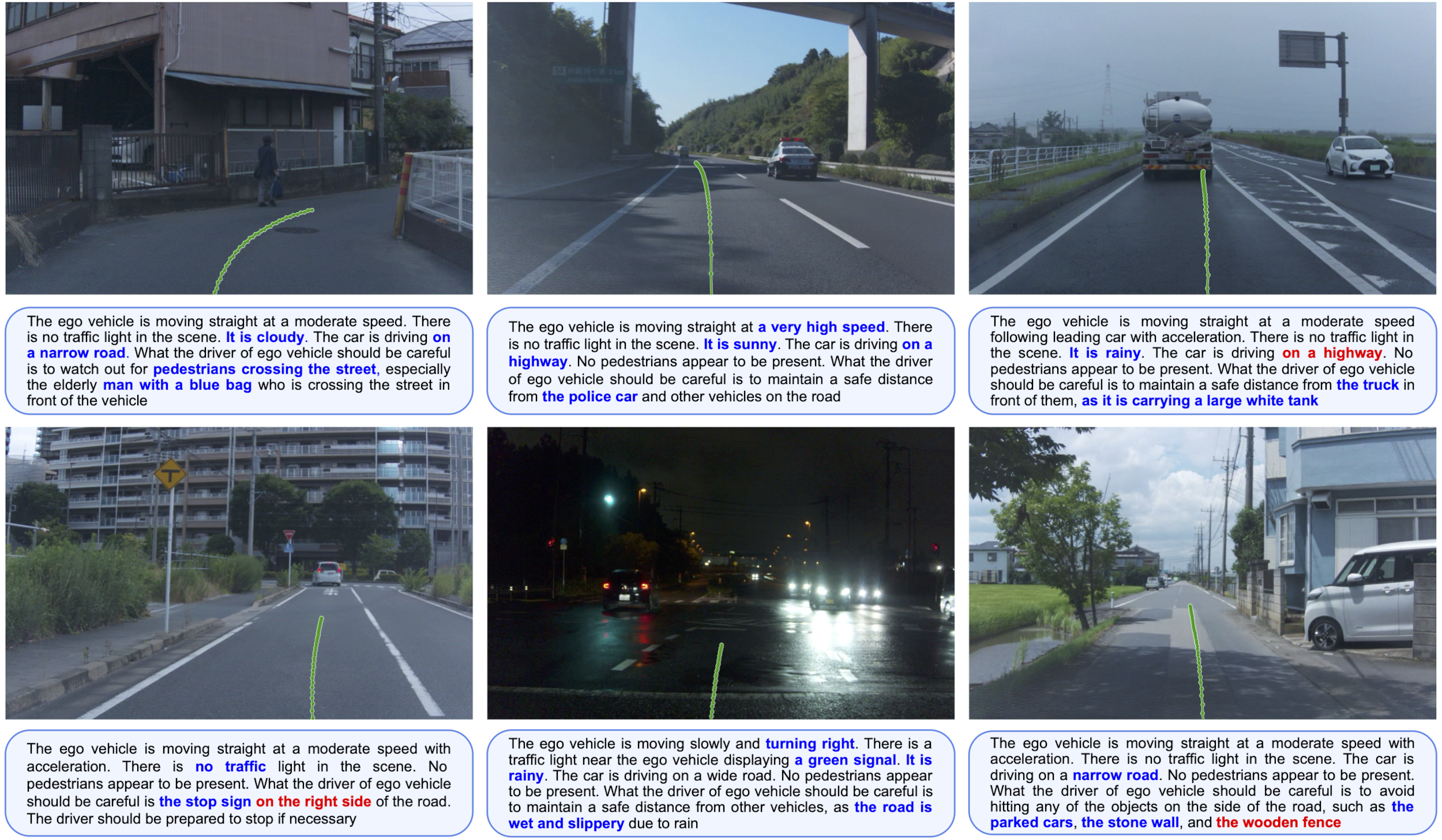}
    \caption{\textbf{Frame examples from CoVLA-Dataset}. Estimated trajectories (\textbf{\textcolor[HTML]{006400}{green line}}) and captions generated by the captioner model are shown. The \textbf{\textcolor{blue}{key objects}} are highlighted in blue bold text, while the \textbf{\textcolor{red}{failures}} in captions are shown in red bold text.}
    \label{fig:dataset_samples}
\end{figure*}

\paragraph{Trajectory}
To acquire future trajectory data, which will be predicted by the VLA models, we estimate the vehicle's travel path with a Kalman Filter~\cite{kalman1960new} using GNSS and IMU sensors.
For each timestamp, we annotate future trajectory data points for the subsequent three seconds (60 frames).
The trajectory data is expressed in the global coordinate system centered on the data collection vehicle.
In some instances, GNSS data instability occasionally leads to erroneous trajectories characterized by significant vibrations. We implement a heuristic approach to identify and remove these inaccurate trajectories from the dataset. Further details are reported in the supplementary material.

\paragraph{Objects}
Traffic lights are among the most critical objects in the driving environment, and accurate detection of their state provides strong support for the training of VLA models. 
We employ a deep learning model specifically designed for traffic light detection (OpenLenda-s\footnote{\url{https://github.com/turingmotors/openlenda/releases/tag/v0.1.0}}) to annotate our data.
This model can accurately identify not only the color of traffic lights but also the direction of arrow signals, providing a comprehensive understanding of traffic light information for each frame.

In addition, accurate detection and tracking of leading vehicles are crucial for understanding traffic flow and predicting potential hazards, making it another vital element of our dataset. To ensure robustness, we employ a sensor fusion approach, combining data from both radar and a front-facing camera. Our dataset includes comprehensive information for the detected leading vehicle, including its speed, acceleration, and position relative to the ego vehicle. 

\subsubsection{Auto Captioning}
\label{sec: auto_captioning}
\paragraph{Rule-based Captioning}
Captions are essential for vision-language-action datasets, but manual annotation presents significant challenges. Human annotation is costly, time-consuming, and often suffers from inconsistencies in caption quality. To address these limitations and ensure scalability, we develop an automated captioning approach.

Rule-based captions in natural language are generated as the first step of our approach. 
We consider various aspects of vehicle motion and detected objects, including speed, acceleration, trajectory curvature, leading vehicle presence, and traffic light status, to generate comprehensive rule-based captions for each frame in the dataset.

\paragraph{VLM-based Captioning}
While efficient and cost-effective, rule-based captioning often lacks the richness of natural language and may overlook critical details such as specific signs or uncommon objects.
To enhance the expressiveness and informativeness of captions, we employ a pretrained VLM to augment the rule-based captions.
While some datasets, such as OpenDV-2K~\cite{yang2024genad}, utilize single-frame auto captioning with models like BLIP-2~\cite{pmlr-v202-li23q}, this approach fails to capture crucial temporal information inherent in driving scenarios.
To address this limitation, we employ a pretrained VideoLLaMA 2\footnote{\url{https://huggingface.co/DAMO-NLP-SG/VideoLLaMA2-7B}}, a large video language model proficient in spatiotemporal modeling and demonstrating strong performance in video question answering and captioning tasks~\cite{cheng2024videollama}.

The captioner model operates on a 60-frame (three-second) window, processing eight representative frames, including the first and last frames of the window, sampled from the input video.
To manage the processing load, each 30-second scene is divided into ten windows. This approach results in 100,000 VLM-generated captions and 6,000,000 combined captions after integrating both rule-based and VLM-generated captions. Caption generation is performed using eight NVIDIA H100 GPUs, with the entire process completes within a single day, demonstrating significant time savings compared to manual annotation.

\paragraph{Hallucination Mitigation}
A known challenge with VLMs is their tendency to generate hallucinations, producing captions that do not accurately reflect the visual content, particularly in tasks like visual question answering and image captioning~\cite{liu2024surveyhallucinationlargevisionlanguage, bai2024hallucination}.

To mitigate this issue, we use comprehensive rule-based captions as factual constraints. These captions are then provided as context to the VLM, along with a prompt instructing it to supplement the captions with any additional information not already covered by the rules.
This supplementary information, derived by querying the model and examining its internal token probability distributions, includes details regarding road type (e.g., narrow/wide, highway/non-highway, tunnel/non-tunnel), weather conditions (e.g., sunny, cloudy, rainy), potential risks (e.g., pedestrian presence), and other relevant factors.
For instance, to determine the weather, we calculate the probability of each weather-related token (e.g., ``sunny,'' ``cloudy,'' or ``rainy'') given the query ``What is the weather in this video?'' and select the token with the highest probability.
Using these augmented rule-based captions as factual anchors, we prompt the VLM to generate free-form captions, focusing on identifying potential risks within the driving environment.

\subsection{Analysis}

\begin{figure}[t!]
    \centering
    \begin{subfigure}[b]{0.45\textwidth}
        \centering
        \includegraphics[width=\textwidth]{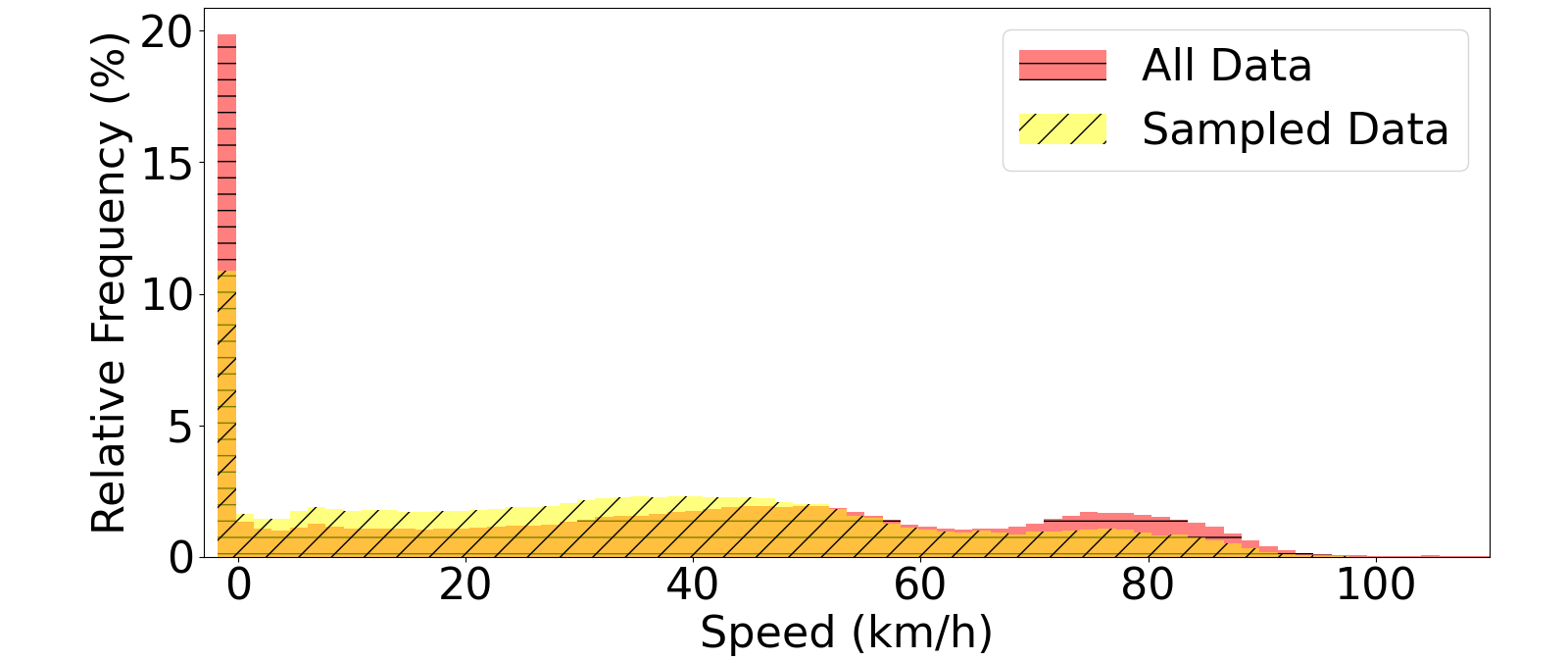}
        \caption{Speed distribution before and after sampling.}
        \label{fig:speed}
    \end{subfigure}
    \hfill
    \begin{subfigure}[b]{0.45\textwidth}
        \centering
        \includegraphics[width=\textwidth]{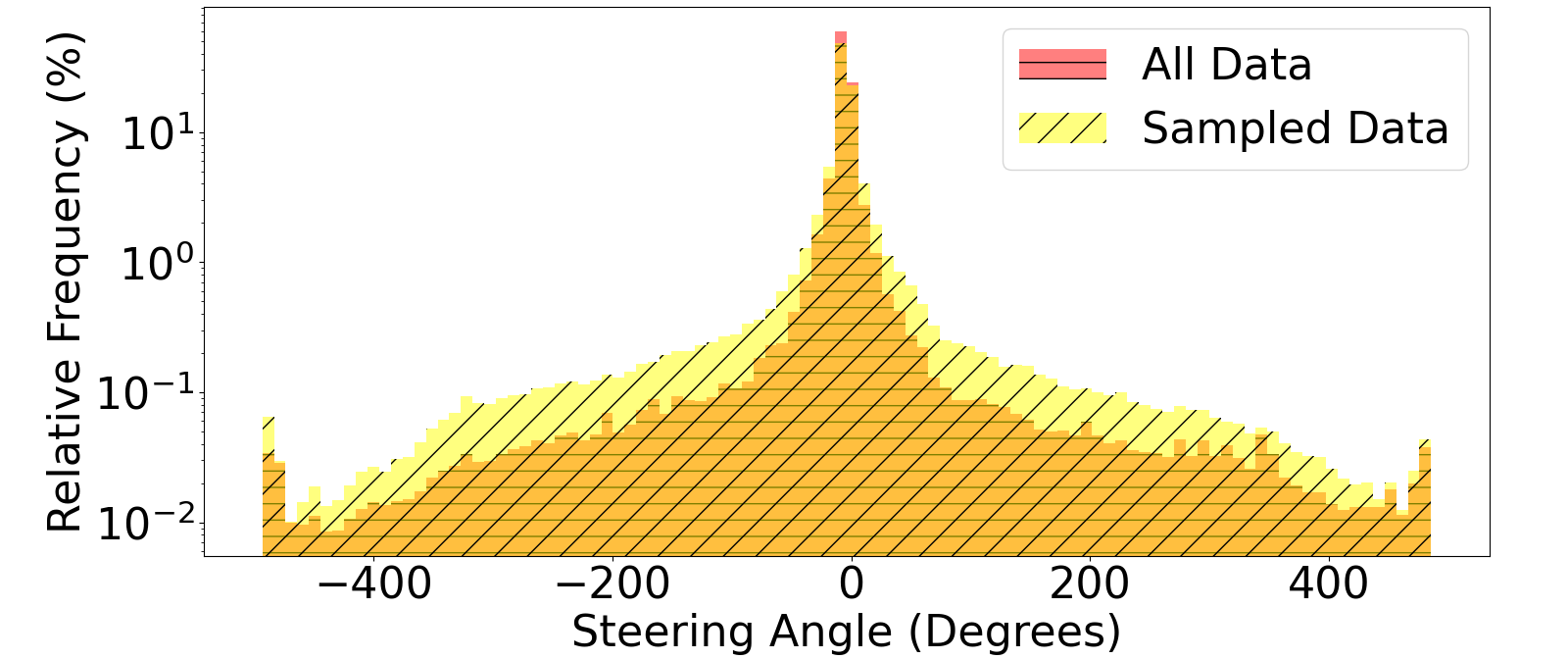}
        \caption{Steering angle distribution before and after sampling.}
        \label{fig:steering_angle}
    \end{subfigure}
    \caption{\textbf{Data distribution of vehicle speed and steering angle}. The red bars represent the distribution before sampling, while the yellow bars show the distribution after sampling. Note that a logarithmic scale is used for clarity in (b).}
\end{figure}

\paragraph{Statistics}
To assess the impact of our scene sampling methodology described in Section \ref{subsubsec:data_sampling}, we first analyze the distribution of key driving parameters.
As shown in \autoref{fig:speed} and \autoref{fig:steering_angle}, the distributions of vehicle speeds and steering angles become more uniform after sampling, with a noticeable increase in lower speed data points and a diminished central peak around zero degrees for steering angles, leading to a more balanced representation across different ranges.
Consequently, CoVLA-Dataset exhibits diverse driving maneuvers, with active turn signals in 16.11\% of frames and traffic lights in 22.90\%. This diversity, including starting/stopping, turning, and lane changing, emphasizes the need for robust perception, decision-making, and control algorithms in autonomous driving systems.

\paragraph{Comparison}
As presented in \autoref{tab:comparison}, our CoVLA-Dataset demonstrates several key advantages. We construct our dataset using auto labeling and MLLMs-based auto captioning similar to OpenDV-2K~\cite{yang2024genad}.
This automated approach is crucial for constructing large-scale datasets, a growing trend in autonomous driving research\cite{li2024data}, and enables us to develop a dataset significantly larger than those reliant on manual annotation.
Furthermore, our dataset incorporates trajectory annotations, a feature absent in datasets lacking sufficient metadata like OpenDV-2K, derived from GPS and IMU data.

CoVLA-Dataset distinguishes itself from other datasets through its seamless integration of vision, language, and action modalities. This integration results from synchronized real-world data acquisition from various sensors, coupled with our automated labeling and captioning pipeline.
This comprehensive annotation captures all relevant aspects of the driving environment, making it a valuable resource for training and evaluating autonomous driving systems.

\paragraph{Limitation of Auto Captioning}
\autoref{fig:dataset_samples} presents frames sampled from CoVLA-Dataset, highlighting both successful examples and instances where auto-captioning encountered challenges.
While our proposed method effectively enables scalable and detailed video captioning, analysis of the generated captions reveals areas for further refinement.

We identify two primary categories of errors. Firstly, the captioning model occasionally exhibits object hallucination, describing objects absent from the scene, such as a ``wooden fence.'' Furthermore, the VLM occasionally misidentifies object locations and scene elements, such as incorrectly labeling left-side objects as ``right-side.''

Another limitation we reveal is the recognition of local landmarks, particularly those unique to Japan. For instance, Japanese traffic signs often have unique designs and meanings, such as including Japanese characters, that necessitate proper identification and interpretation.

These findings highlight key areas for future research, including the development of more robust object detection algorithms, context-aware captioning models with more knowledge of the driving environment, and culturally diverse training datasets.

\section{Experiment}

In this section, we present the development and evaluation methodology of our baseline model, CoVLA-Agent, which leverages the richness of CoVLA-Dataset for autonomous driving tasks. We detail the experimental setup, including dataset, model configurations, training procedures, and evaluation metrics, and analyze the results.
\subsection{Model}
\begin{figure}[ht]
\centering
\includegraphics[width=0.9\linewidth]{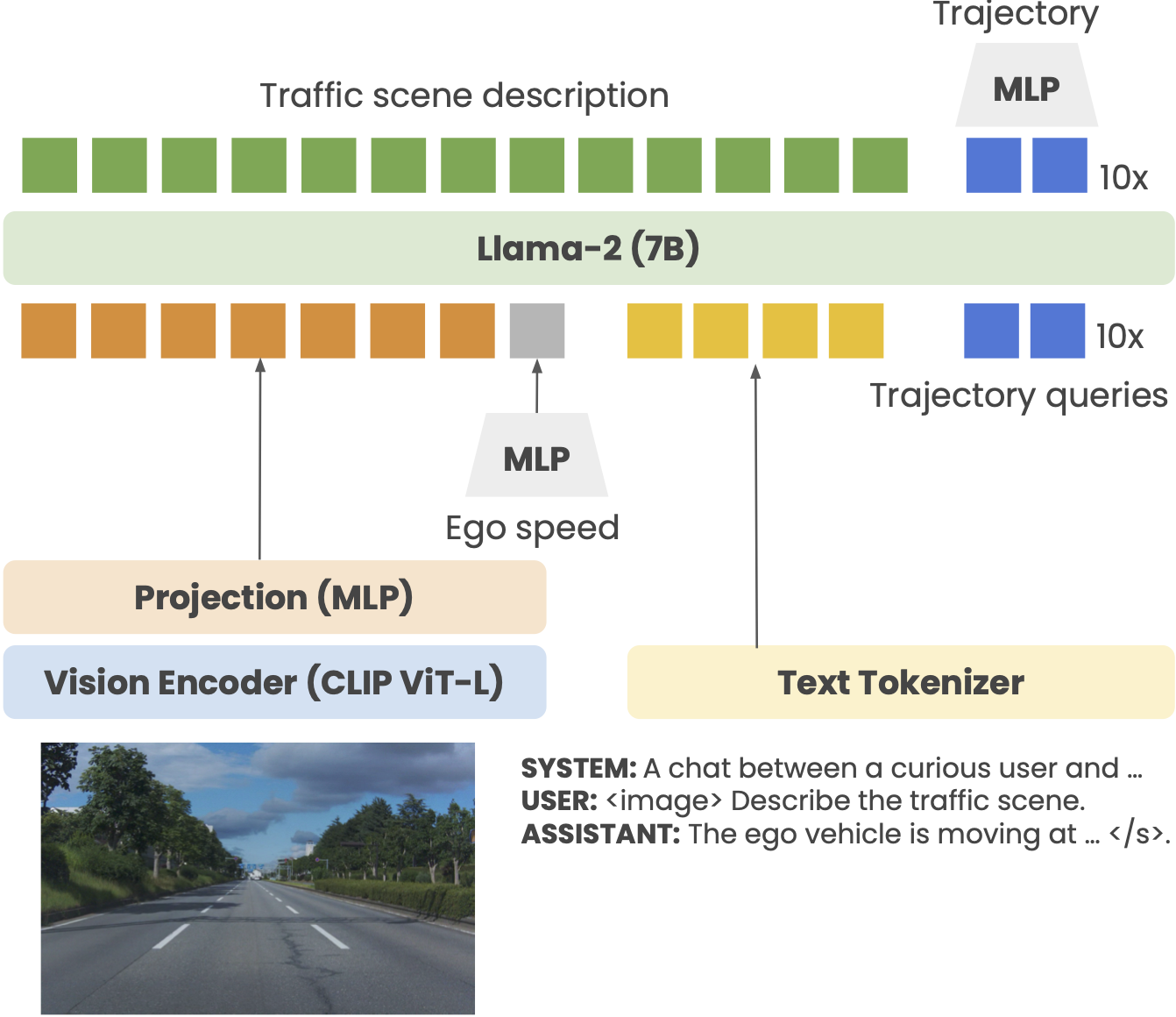}
\caption{\textbf{The architecture for CoVLA-Agent}.}
\label{fig:architecture_covla_agent}
\end{figure}

\paragraph{Architecture}
CoVLA-Agent as illustrated in \autoref{fig:architecture_covla_agent}, is a VLA model designed for autonomous driving. We utilize the pretrained Llama-2 (7B)~\cite{touvron2023llama} as a language model and CLIP ViT-L (224$\times$224 pixels)~\cite{pmlr-v139-radford21a} as a vision encoder.
In addition, our model takes the ego vehicle's speed as an input, which is transformed into an embedding vector using a Multi Layer Perceptron (MLP).
The visual features extracted by CLIP ViT-L are concatenated with the speed embedding and text embedding, and then fed into the Llama-2 model.
For trajectory prediction, special tokens are used as trajectory queries. 
The outputs of these trajectory queries are processed by an MLP layer, resulting in a sequence of 10 $(x,y,z)$ coordinates representing the predicted trajectory of the vehicle relative to its current position, spanning a three-second time horizon.

\paragraph{Training}
Based on this architecture, we train CoVLA-Agent on two tasks, which are traffic scene description generation and trajectory prediction.
For traffic scene description generation, we use cross entropy loss as the loss function, and for trajectory prediction, we employ mean squared error loss. Ultimately, the training aims to minimize a combined loss function where both losses are weighted equally. 

\subsection{Experimental Settings}

\paragraph{Data}
CoVLA-Agent is trained on CoVLA-Dataset.  We split the 10,000 scenes in the dataset into a 70/15/15 for training, validation, and testing, respectively.
For each scene, we sample frames at a frequency of 2Hz, excluding any frames that do not contain all trajectory coordinate points for the subsequent three seconds (60 coordinate points in total). From these 60 coordinate points, we uniformly sample 10 points to represent the future trajectory. This process yields a data subset comprising 302,989 samples for training, 64,153 samples for validation, and 64,920 samples for testing.
We preprocess the training and validation datasets into the LLaVA instruction tuning format \cite{NEURIPS2023_6dcf277e} for efficient model training. This format, illustrated in \autoref{fig:architecture_covla_agent}, consists of a system prompt, a user instruction, and an assistant response.

\paragraph{Conditions}
To examine the impact of the captioning task on subsequent trajectory prediction, we prepare two conditions: the \textit{predicted caption} condition, where CoVLA-Agent generates captions prior to predict trajectory, and the \textit{ground truth caption} condition, where actual captions are used instead of predicted ones during trajectory prediction.

\paragraph{Metrics}
We compute both Average Displacement Error (ADE) and Final Displacement Error (FDE) to assess the trajectory prediction accuracy~\cite{phong2024truly}. ADE is computed as the mean Euclidean distance between the predicted trajectory points and the ground truth trajectory points over all time steps. FDE measures the Euclidean distance between the predicted final point and the ground truth final point.

\begin{displaymath}
\begin{aligned}
\text{ADE} &= \frac{1}{T} \sum_{t=1}^{T} \sqrt{(x_t - \hat{x}_t)^2+ (y_t - \hat{y}_t)^2 + (z_t - \hat{z}_t)^2} \\
\text{FDE} &= \sqrt{(x_T - \hat{x}_T)^2 + (y_T - \hat{y}_T)^2  + (z_T - \hat{z}_T)^2}
\end{aligned}
\end{displaymath}

\subsection{Results}
We present qualitative results in \autoref{tab:results}. The ground truth captions achieve a lower ADE (0.814) and FDE (1.655) compared to the predicted captions’ ADE (0.955) and FDE (2.239). The results indicate that CoVLA-Agent performs better when using ground truth captions compared to predicted captions. 

\newcolumntype{F}{>{\centering\arraybackslash}p{1.8cm}}
\newcolumntype{Y}{>{\centering\arraybackslash}p{1.5cm}}
\newcolumntype{Z}{>{\centering\arraybackslash}p{1.5cm}}

\begin{table}[h]
\centering
\scalebox{0.85}{
\begin{tabularx}{\linewidth}{XFF}
\toprule
\multirow{2}{*}{\textbf{Condition}} & \multicolumn{2}{c}{\textbf{Metrics}} \\
\cmidrule(lr){2-3}
& \textbf{ADE $\downarrow$} & \textbf{FDE $\downarrow$} \\
\midrule
predicted caption    & 0.955 & 2.239\\ 
ground truth caption & \textbf{0.814} & \textbf{1.655}\\ 
\bottomrule
\end{tabularx}
}
\caption{\textbf{Quantitative comparison of different conditions}.}
\label{tab:results}
\end{table}

\subsection{Discussion}

\begin{figure}[b]
\centering
\begin{subfigure}[b]{0.48\linewidth}
    \centering
    \includegraphics[width=\linewidth]{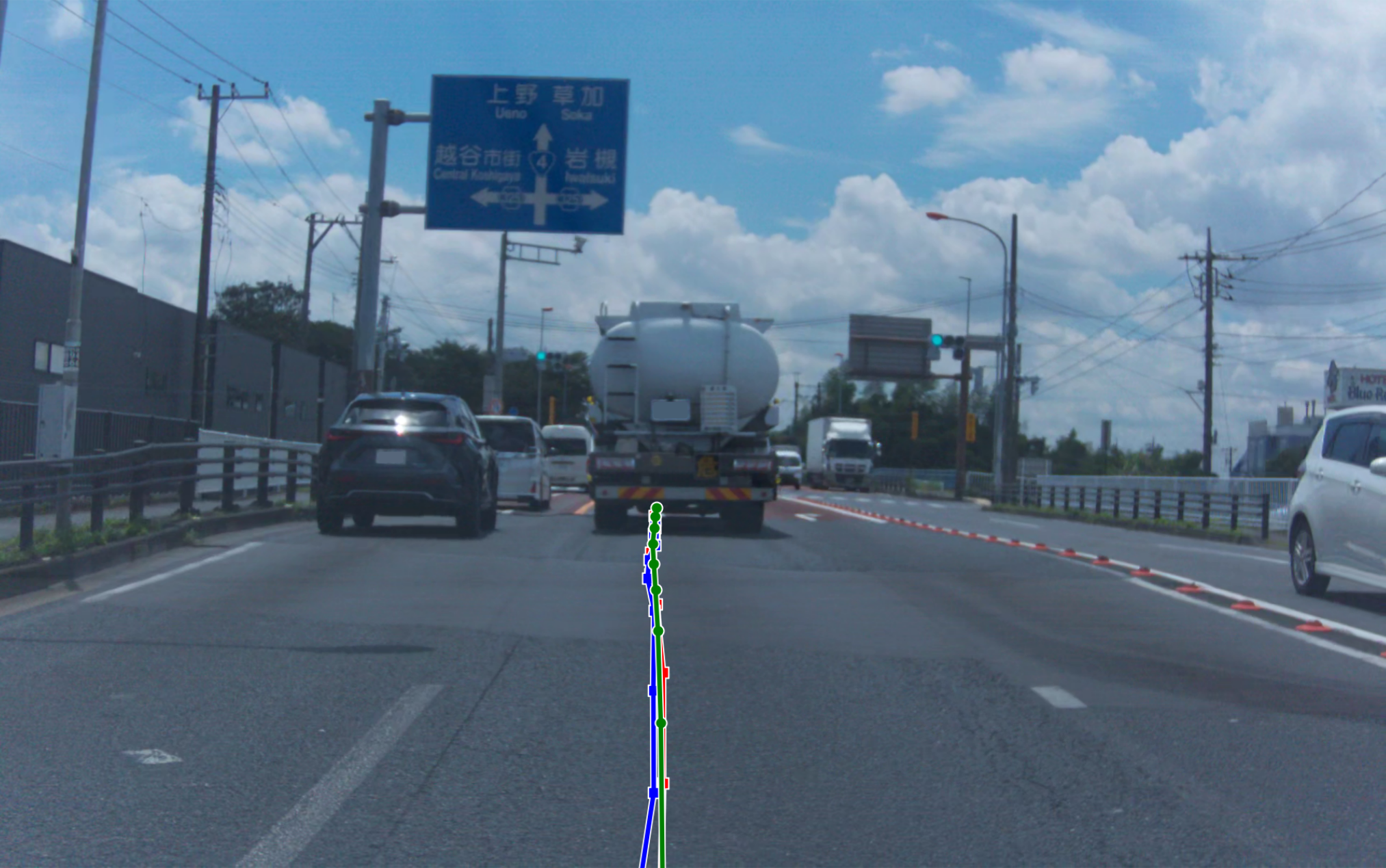}
    \caption{A scene on a local road with three lanes each way. There is a silver tank truck ahead.}
    \label{fig:6(a)}
\end{subfigure}
\hfill
\begin{subfigure}[b]{0.48\linewidth}
    \centering
    \includegraphics[width=\linewidth]{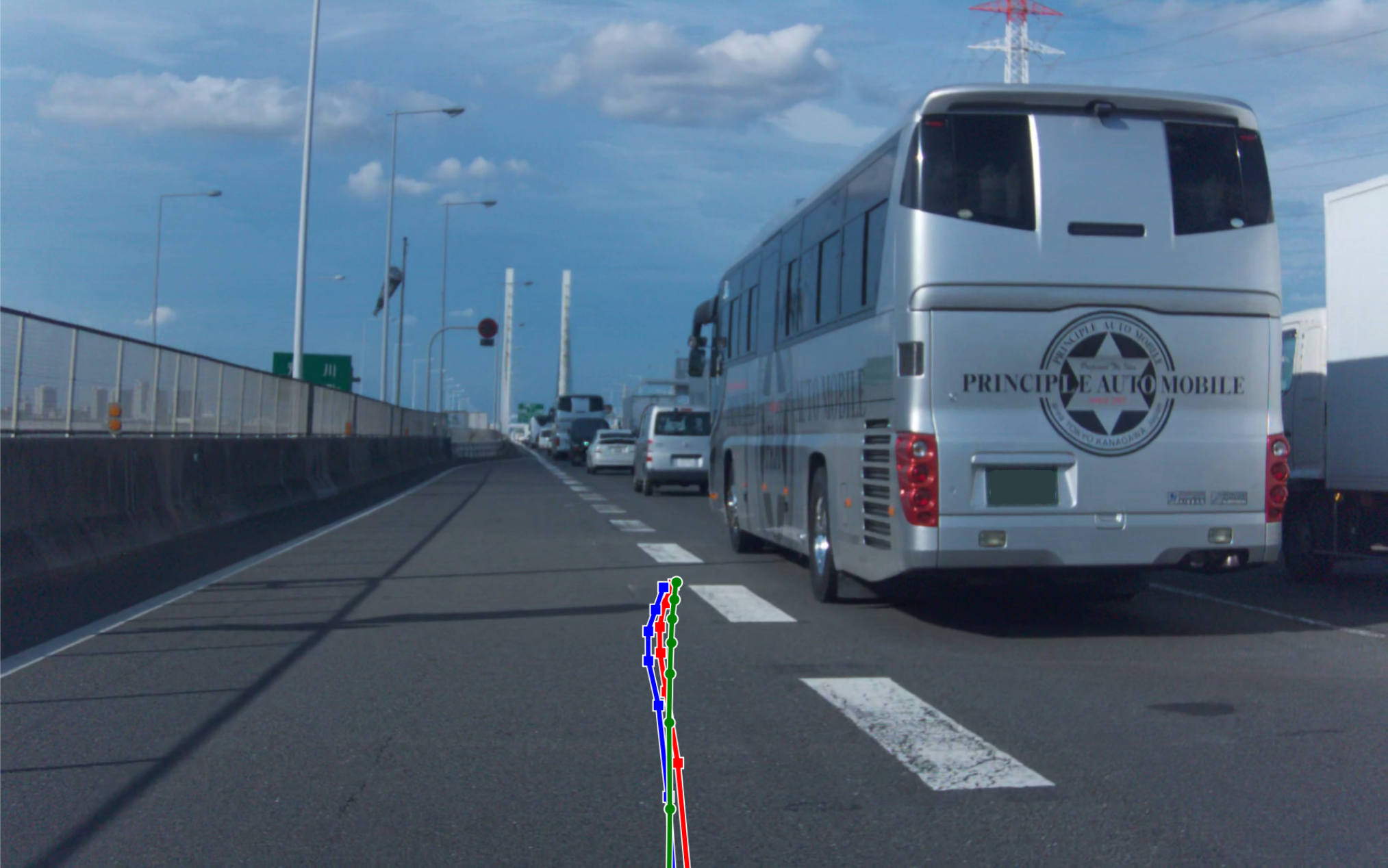}
    \caption{A scene on a freeway. The ego vehicle is about to merge behind a silver bus on the main road.}
    \label{fig:6(b)}
\end{subfigure}
\vspace{0.5cm}
\begin{subfigure}[b]{0.48\linewidth}
    \centering
    \includegraphics[width=\linewidth]{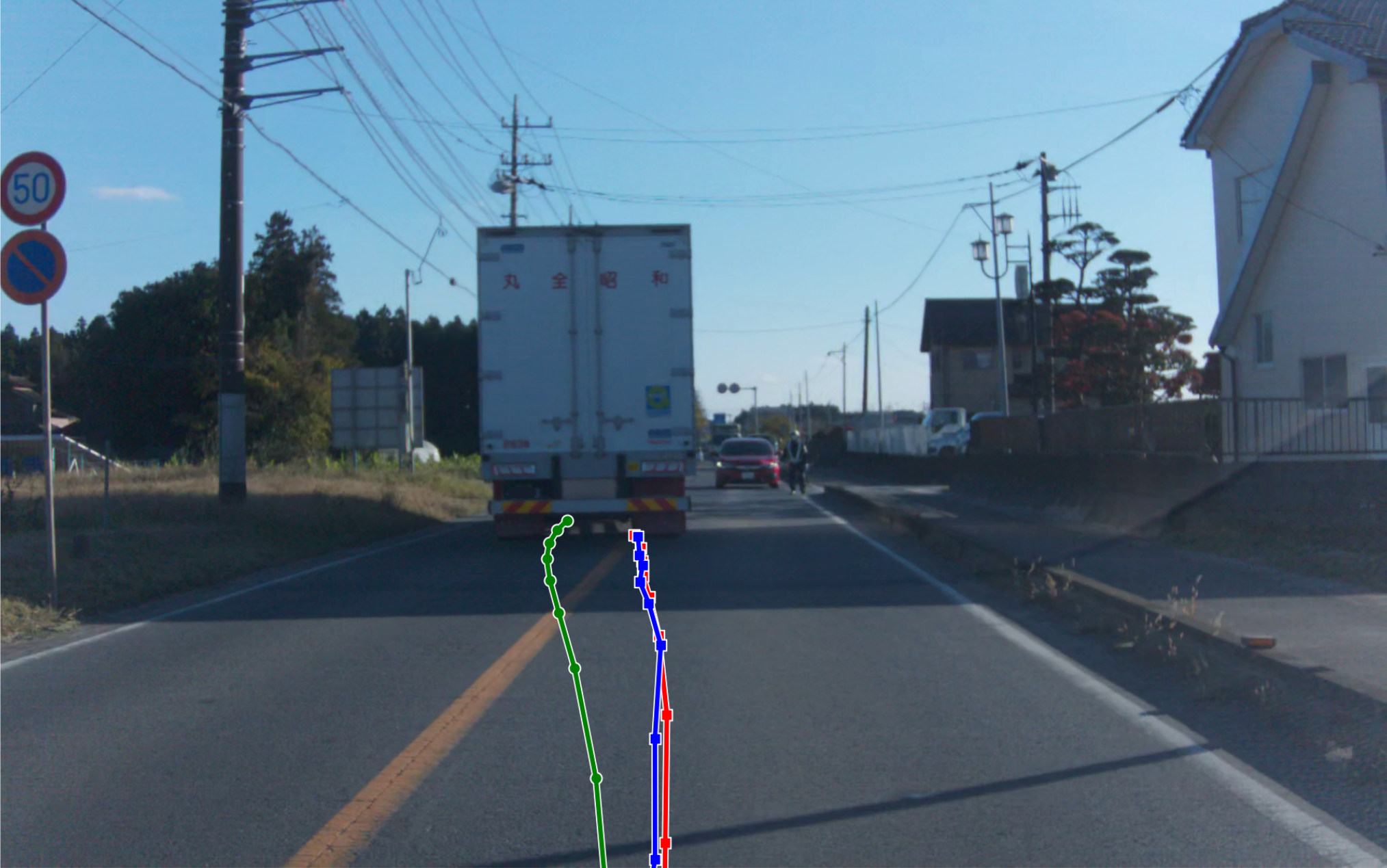}
    \caption{A scene of traffic control. The ego vehicle is in the right lane and needs to return to the left lane.}
    \label{fig:6(c)}
\end{subfigure}
\hfill
\begin{subfigure}[b]{0.48\linewidth}
    \centering
    \includegraphics[width=\linewidth]{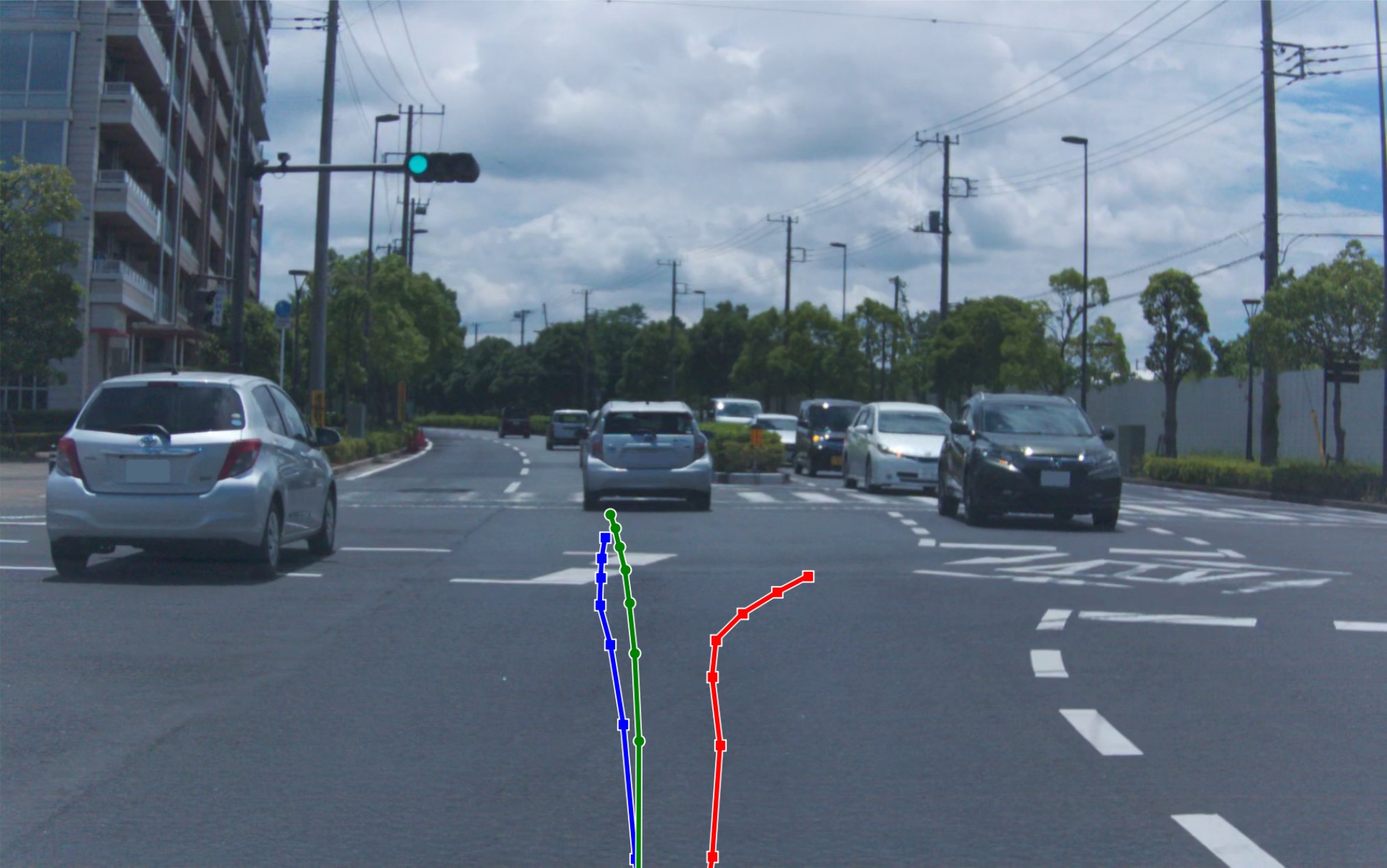}
    \caption{A scene at an intersection. The traffic light is green. There are several cars lined up in the right lane.}
    \label{fig:6(d)}
\end{subfigure}
\caption{\textbf{Results of trajectory prediction by CoVLA-Agent across various traffic scenes}. \textbf{\textcolor{red}{Red line}} is the predicted trajectory with predicted caption condition, \textbf{\textcolor{blue}{blue line}} is the trajectory predicted with ground truth caption condition and \textbf{\textcolor[HTML]{006400}{green line}} is the ground truth trajectory.}
\label{fig:6}
\end{figure}
We analyze our model's performance across various driving scenarios.

\autoref{fig:6(a)} demonstrates accurate trajectory prediction on a straight road, while \autoref{fig:6(b)} showcases successful freeway merging. \autoref{fig:6(c)} illustrates a scene where the ego vehicle is in the right lane due to traffic control. A traffic officer stops a red car ahead in the right lane. To avoid them, our model predicts a trajectory that returns to the left lane. Our approach is capable of producing accurate routes even in such complex traffic control scene.

\autoref{fig:6(d)} shows the results of trajectory prediction at an intersection. In this scene, the ground truth caption includes the phrase ``\textit{moving straight}'' and the trajectory predicted with that condition is straight. In contrast, the predicted trajectory based on the predicted caption including phrase ``\textit{turning right},'' accurately reflects a right-turn trajectory. While this result indicates the difficulty of estimating intention from a single frame in such scenes, it also underscores the consistency between language and action modality in CoVLA-Agent.

To examine the impact of these captioning errors on trajectory, we investigate how errors in predicted captions affect ADE and FDE. Specifically, we analyze the mean ADE and FDE values for each word in the captions when the word appears in the ground truth rule-based caption part but not in the predicted caption, and vice versa. Excluding stop words, the top ten words with frequencies exceeding ten that led to the largest ADE and FDE values are shown in \autoref{table:ade_fde_for_each_word}, though their rankings vary between the two metrics. Large ADE and FDE occur when there are discrepancies in the predicted caption, particularly with words related to vehicle motion direction and acceleration, such as ``deceleration,'' ``left,'' ``curve,'' and ``turning.'' Based on these observations, we attribute the relatively low performance in the trajectory predicted using the generated caption largely to the difficulty of estimating intention from a single frame.

\begin{table}[h]
\setlength\dashlinedash{0.5pt}
\setlength\dashlinegap{2.0pt}
\setlength{\tabcolsep}{3.6pt}

\centering
\scalebox{0.90}{
\begin{tabular}{l|ccc}
\hline
\textbf{Word} & \textbf{Mean ADE $\downarrow$} & \textbf{Mean FDE $\downarrow$} & \textbf{Frequency} \\ \hline\hline
\textbf{deceleration} & 2.236 & 5.458 & 324 \\ \hdashline
\textbf{left} & 2.037 & 5.009 & 274\\ \hdashline
\textbf{acceleration} & 1.826 & 4.790 & 50 \\ \hdashline
\textbf{curve} & 1.527 & 3.608 & 167 \\ \hdashline
following & 1.400 & 3.192 & 20 \\ \hdashline
\textbf{turning} & 1.343 & 3.288 & 636 \\ \hdashline
\textbf{right} & 1.324 & 3.282 & 810\\ \hdashline
appear & 1.244 & 3.044 & 93 \\ \hdashline
present & 1.244 & 3.044 & 93 \\ \hdashline
signal & 1.229 & 3.049 & 117 \\ \hline
\end{tabular}
}
\caption{\textbf{Top-10 words with the largest mean ADE and FDE}. These words correspond to motions that are difficult to estimate from a single frame. Words explicitly signifying motion are presented in bold.}
\label{table:ade_fde_for_each_word}
\end{table}

These behaviors demonstrate that our CoVLA-Dataset led the model to generate consistent captions and trajectories, showing an outstanding capability to handle a variety of driving scenes and is useful for training autonomous driving models to manage complex scenarios. CoVLA-Dataset accelerates future research on the application of VLA models for self-driving vehicles.

\section{Conclusion}
In this study, we presented CoVLA-Dataset, a novel dataset for autonomous driving with VLA models. By leveraging scalable automated approach, we have constructed a large-scale, comprehensive dataset enriched with detailed language annotations. Building on this robust dataset, we developed CoVLA-Agent, a sophisticated VLA autonomous driving model. The evaluation results underscore the strong capability of the model which generates coherent language and action outputs. These findings highlight the transformative potential of VLA multi-modal models and pave the way for future innovations in autonomous driving research.

\section*{Acknowledgments}
This paper is based on results obtained from a project, JPNP20017, subsidized by the New Energy and Industrial Technology Development Organization (NEDO).

{\small
\bibliographystyle{ieee_fullname}
\bibliography{egbib}
}
\newpage

\appendix
\onecolumn
\begin{center}
    {\large \textbf{CoVLA: Comprehensive Vision-Language-Action Dataset for Autonomous Driving \\--- Supplementary Material ---}}
\end{center}

\section{Heuristic Trajectory Filtering}
In some instances, trajectory data instability was detected. Specifically, we identified two erroneous behaviors:
\begin{enumerate}
    \item Significant jumps
    \item Movement in the wrong direction
\end{enumerate}
To detect significant jumps, we filtered the trajectory data by the distance between adjacent points. Given a recording frequency of 20 Hz and a maximum speed of 100 km/h, the distance between points should be at most 1.38 meters, which is calculated as following:

\begin{equation}
\frac{100 \, \text{km/h} \times 1000 \, \text{m/km}}{3600 \, \text{s/h} \times 20} = 1.38 \, \text{m}
\end{equation}

With a tolerance rate of 1.15, the threshold was set to 1.59 meters. All trajectories exceeding this threshold were filtered.

To detect movement in the wrong direction, we manually checked 400 samples from all scenes and identified 43 invalid trajectories (10.75\%). Observations revealed a vibration frequency of 10 Hz in these trajectories. To implement vibration detection, we smoothed the trajectory using a 3-point moving average and calculated the difference between the smoothed trajectory and the original trajectory. We then analyzed the variance of these differences. If the variance exceeded a certain threshold, the trajectory was classified as invalid. This method yielded a precision of 0.64 and a recall of 0.75 on the test dataset, reducing the invalid trajectory rate to 2.6\%. Although this method has a relatively high false-positive rate, it is acceptable for the dataset's scale.

\section{Privacy Protection for Dataset Publication}

We published CoVLA-Dataset~\footnote{\url{https://huggingface.co/datasets/turing-motors/CoVLA-Dataset}} on HuggingFace. For privacy protection, we anonymized human faces and license plates in the CoVLA-Dataset images and videos, using Dashcam Anonymizer~\footnote{\url{https://github.com/varungupta31/dashcam_anonymizer}}.

\section{Example data in CoVLA-Dataset}

\begin{figure}[ht]
\centering
\includegraphics[width=0.39\linewidth]{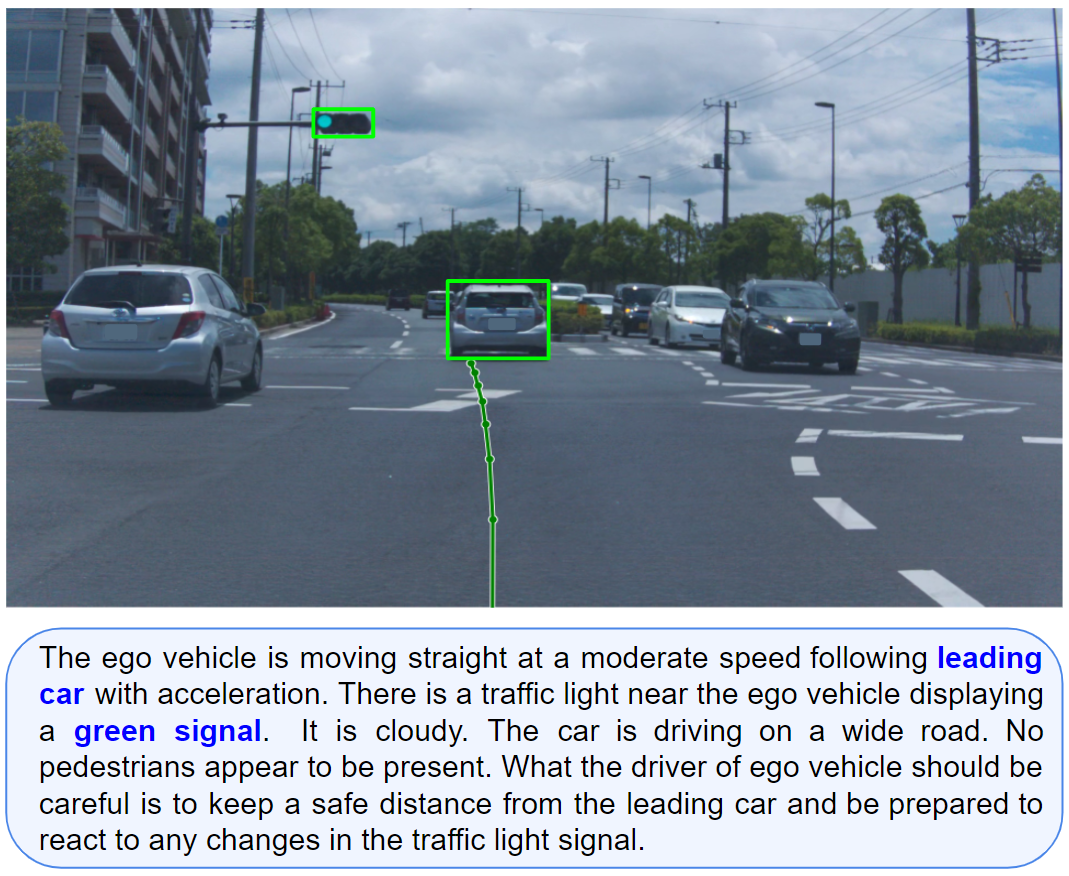}
\caption{Example of image and caption in CoVLA-Dataset.}
\label{fig:7}
\end{figure}

We present sample data from CoVLA-Dataset. Each data point includes an image, a caption, and vehicle states. An example of an actual caption is shown in \autoref{fig:7}. For auto-captioning, we use the traffic light detection provided by OpenLenda-s and the front car detection results from the sensor fusion. The list of vehicle states is indicated in \autoref{table:trajectory_data}. Each frame contains information on the speed, steering angle, coordinates of the trajectory, and more.

\begin{table*}[h!]
    \centering
    \scriptsize
    \renewcommand{\arraystretch}{2.0}
    \begin{tabular}{|>{\raggedright\arraybackslash}p{3.0cm}|>{\raggedright\arraybackslash}p{14.0cm}|}
        \hline
        \textbf{Key} & \textbf{Value} \\
        \hline
        frame\_id & 569 \\
        image\_path & images/2022-07-08--11-37-27--5\_first/0569.png \\
        vEgo & 7.43082332611084 \\
        vEgoRaw & 7.4395833015441895 \\
        aEgo & 0.6044138669967651 \\
        steeringAngleDeg & 0.6073870658874512 \\
        steeringTorque & 69.0 \\
        brake & 0.0 \\
        brakePressed & false \\
        gas & 0.20499999821186066 \\
        gasPressed & true \\
        doorOpen & false \\
        seatbeltUnlatched & false \\
        gearShifter & drive \\
        leftBlinker & false \\
        rightBlinker & false \\
        orientations\_calib & \(\begin{bmatrix} 2.9467956252753775 & 0.9174552319868815 & 2.2181786819453384 \end{bmatrix}\) \\
        orientations\_ecef & \(\begin{bmatrix} 2.9243210649189977 & 0.9224135550861058 & 2.1900513923432348 \end{bmatrix}\) \\
        orientations\_ned & \(\begin{bmatrix} -0.013463193567253392 & 0.006326533926443111 & -2.990125370637735 \end{bmatrix}\) \\
        positions\_ecef & \(\begin{bmatrix} -3959574.486029379 & 3328427.354910454 & 3719065.7393601397 \end{bmatrix}\) \\
        velocities\_calib & \(\begin{bmatrix} 7.317097759615114 & 0.003242519329502727 & 0.005369323447773883 \end{bmatrix}\) \\
        velocities\_ecef & \(\begin{bmatrix} -2.6767882666706004 & 3.547338396353873 & -5.813015899212604 \end{bmatrix}\) \\
        accelerations\_calib & \(\begin{bmatrix} 0.4734579094803297 & 0.08559864698994124 & -0.13132037594775653 \end{bmatrix}\) \\
        accelerations\_device & \(\begin{bmatrix} 0.4736293760658642 & 0.07819260264673351 & -0.13526157094253702 \end{bmatrix}\) \\
        angular\_velocities\_calib & \(\begin{bmatrix} 0.011550541795216845 & 0.012243857869171634 & -0.007753300486330907 \end{bmatrix}\) \\
        angular\_velocities\_device & \(\begin{bmatrix} 0.011675888068301523 & 0.01206267096884485 & -0.007848971055415363 \end{bmatrix}\) \\
        timestamp & 1657248173200 \\
        extrinsic\_matrix & \(\begin{bmatrix} 
            -0.015688330416257182 & -0.9998769191404183 & 0.00012959444326649344 & 0.0 \\ 
            -0.008260370686184616 & 2.879912020664621e-21 & -0.9999658837914467 & 1.2200000286102295 \\ 
            0.9998428078989188 & -0.01568886620613436 & -0.008259354077745229 & 0.0 \\ 
            0.0 & 0.0 & 0.0 & 1.0 
        \end{bmatrix}\) \\
        intrinsic\_matrix & \(\begin{bmatrix} 
            2648.0 & 0.0 & 964.0 \\ 
            0.0 & 2648.0 & 604.0 \\ 
            0.0 & 0.0 & 1.0 
        \end{bmatrix}\) \\
        trajectory\_count & 60 \\
        \hline
    \end{tabular}
    \label{table:vehicle_data}
\end{table*}

\newpage

\begin{table*}[t]
    \centering
    \scriptsize
    \renewcommand{\arraystretch}{2.0}
    \begin{tabular}{|>{\raggedright\arraybackslash}p{3.0cm}|>{\raggedright\arraybackslash}p{14.0cm}|}
         \hline
        trajectory & \(\begin{bmatrix} 
            0.0 & -0.0 & 0.0 \\ 
            ... & ... & ... \\
            2.2367286927600376 & 0.021314714278469867 & 0.011369742248091336 \\ 
            ... & ... & ... \\
            4.529136516056675 & 0.05988551136966269 & 0.03572205827208557 \\ 
            ... & ... & ... \\
            6.919731941586808 & 0.08958899605082428 & 0.050459818682157966 \\ 
            ... & ... & ... \\
            9.301852576186251 & 0.1417775525229876 & 0.07520389298492622 \\ 
            ... & ... & ... \\
            11.677091075613516 & 0.2108797042962825 & 0.09371664992653018 \\ 
            ... & ... & ... \\
            14.027571172292527 & 0.28508464282229706 & 0.10760938523286476 \\
            ... & ... & ... \\
            16.350994760951718 & 0.38061347529509826 & 0.1208462683905208 \\ 
            ... & ... & ... \\
            18.675065252972097 & 0.4777731491748536 & 0.14622174266379429 \\ 
            ... & ... & ... \\
            20.998125620182435 & 0.5860796022647292 & 0.1657656398148593 
        \end{bmatrix}\) \\
        \hline
    \end{tabular}
    \caption{Example of Vehicle States List.}
    \label{table:trajectory_data}
\end{table*}

\end{document}